%% file: acl_main.tex
\pdfoutput=1

\documentclass[11pt]{article}
\usepackage[final]{ACL2023}
\usepackage{times}
\usepackage{latexsym}
\usepackage{amsmath}
\usepackage{pifont}
\usepackage{graphicx}
\usepackage{subcaption}
\usepackage{longtable}
\usepackage{float}
\usepackage{arydshln} 
\usepackage{booktabs}
\usepackage{amssymb,amsmath,amsthm}
\usepackage{adjustbox}
\usepackage{multirow}
\usepackage{tabularx} 
\usepackage{tipa} 
\usepackage{fancyvrb}
\usepackage{tcolorbox}
\usepackage{enumitem}
\usepackage{needspace}
\usepackage[T1]{fontenc}
\usepackage[utf8]{inputenc}
\usepackage{microtype}
\usepackage{inconsolata}
\usepackage{arydshln}
\usepackage{colortbl} % Include this in the preamble
\usepackage{cleveref}
\crefname{section}{Sec}{Secs}
\crefname{figure}{Fig}{Figs}
\crefname{table}{Tab}{Tabs}
\crefname{equation}{Eq}{Eqs}
\crefname{appendix}{Appendix}{Appendices}

\title{ReSCORE: Label-free Iterative Retriever Training for Multi-hop Question Answering with Relevance-Consistency Supervision}

\author{%
Dosung Lee$^{1*}$ \quad Wonjun Oh$^{1}$\thanks{* Equal contribution.} \quad Boyoung Kim$^{1}$ \quad Minyoung Kim$^{1}$ \\ \textbf{Joonsuk Park}$^{2,3,4}$\textsuperscript{\(\dagger\)} \quad \textbf{Paul Hongsuck Seo}$^{1}$\thanks{\(\dagger\) Corresponding authors.} 
\\
$^{1}$Dept. of CSE, 
Korea University,\\
  $^{2}$NAVER AI Lab, 
  $^{3}$NAVER Cloud, \\
  $^{4}$University of Richmond \\
  {\small{\texttt{\{dslee1219, owj0421, bykimby, omniverse186, phseo\}@korea.ac.kr}}} \\
  \small{\texttt{{park@joonsuk.org} \quad }}}
  
\begin{document}
\maketitle
\begin{abstract}
Multi-hop question answering (MHQA) involves reasoning across multiple documents to answer complex questions. Dense retrievers typically outperform sparse methods like BM25 by leveraging semantic embeddings; however, they require labeled query-document pairs for fine-tuning. This poses a significant challenge in MHQA due to the high variability of queries---(reformulated) questions---throughout the reasoning steps. To overcome this limitation, we introduce Retriever Supervision with Consistency and Relevance (ReSCORE), a novel method for training dense retrievers for MHQA without labeled documents. ReSCORE leverages large language models to capture each document's relevance to the question and consistency with the correct answer and use them to train a retriever within an iterative question-answering framework. Experiments on three MHQA benchmarks demonstrate the effectiveness of ReSCORE, with significant improvements in retrieval, and in turn, the state-of-the-art MHQA performance. Our implementation is available 
% \href{https://github.com/leeds1219/ReSCORE}{here}.
at: \href{https://leeds1219.github.io/ReSCORE}{https://leeds1219.github.io/ReSCORE}.
\end{abstract}

\input{src/01_introduction}

\input{src/02_related_works}
\input{src/03_methods}

\input{src/04_experiments}
\input{src/05_conclusion}

\section*{Limitations}
The fine-tuning process for our model is specifically tuned to datasets such as MuSiQue, 2WikiMultiHopQA, and HotpotQA, each of which has distinct characteristics, including the required number of hops and the types of reasoning involved. 
While our retriever demonstrates strong performance on trained datasets, its ability to generalize to other datasets that differ in reasoning patterns or dataset characteristics remains limited. 
This limitation highlights an Out-of-Distribution generalization challenge.
Also, our approach relies on an iterative retrieval process, which increases computational costs and latency, especially for questions with high hop requirements. In practical applications
% , such as real-time question answering, 
the computational demand may be prohibitive. Further optimization is necessary to make the framework more efficient and scalable.

\section*{Ethics Statement}
This study adheres to ethical standards, emphasizing fairness, transparency, and responsibility. All datasets (MuSiQue, 2WikiMultiHopQA, HotpotQA) are publicly available, curated, and free of personally identifiable information. They are released under the MIT License, permitting modification, redistribution, and use with proper attribution.
The Contriever model is released under the Attribution-NonCommercial 4.0 International (CC BY-NC 4.0) license, allowing non-commercial use, modification, and redistribution with proper attribution.
The META LLAMA 3.1 model is released under the LLAMA 3.1 COMMUNITY LICENSE AGREEMENT (Release Date: July 23, 2024), governing responsible use, modification, and redistribution in accordance with META’s terms.
We ensured consistency in training and evaluation conditions to maintain unbiased comparisons. We recognize the broader implications of multi-hop question-answering advancements and are committed to responsible development and application.

\section*{Acknowledgements}
This research was supported by IITP grants (IITP\allowbreak-2025-RS-2020-II201819, IITP-2025-RS-2024-00436857, IITP-2025-RS-2024-00398115, IITP-2025-RS-2025-02263754, IITP-2025-RS-\allowbreak2025-02304828), the NRF grant (NRF-2021R\allowbreak1A6A1A03045425) and the KOCCA grant (RS-2024-00345025) funded by the Korea government (MSIT, MOE and MSCT).

\bibliography{anthology,custom}
\bibliographystyle{acl_natbib}

\clearpage

\input{src/06_appendix}

\end{document}

%% file: src/01_introduction.tex
\section{Introduction}
\input{figures/01_fig}
Multi-hop question answering (MHQA) consists of complex questions that need to be answered by logically-connecting relevant information from multiple documents.
%subquestions 
% \dslee{this seems like question decompositon more likely, which is one of the way to approach MHQA in certain papers but a broader spectrum including tasks such as longformQA, rather than our iterative approach, most works on MHQA task define MHQA as: extracting and combining multiple pieces of information and doing multiple steps of reasoning \cite{2022arXiv220409140M}, aggregating information from multiple documents, requiring logical reasoning or sequential processing \cite{xiong2020answering} and inference skills by requiring a model to read multiple paragraphs to answer a given question\cite{ho2020constructing}.}
% , often based on information available in multiple documents, 
% and synthesizing the results. 
For instance, to answer "\textit{Which city was the director of the film Parasite born?}", you must first identify the director---"\textit{Bong Joon-ho}"---and figure out where he was born---"\textit{Bongdeok-dong, Daegu}." 
% Recent advancements in MHQA have utilized techniques such as Chain-of-Thought (CoT) reasoning and question decomposition~\cite{trivedi2022interleaving, kim2023tree}. 
% These state-of-the-art methods decompose complex questions into simpler sub-questions or intermediate reasoning steps, enabling models to reason step-by-step and arrive at the final answer.
The state-of-the-art (SOTA) systems for MHQA
take an 
%multi-hop
iterative retrieval-augmented generation (RAG) approach, where they iteratively retrieve relevant documents and generate partial answers from them, until the final answer is reached, as illustrated in Fig.~\ref{fig:inference}~\cite{trivedi2022interleaving,jeong2024adaptive}.
% find answers to the subquestions using retrieval-augmented generation (RAG), where documents (hopefully) containing relevant information are first retrieved and provided to a large language model (LLM) along with the question to generate an answer more reliably~\cite{trivedi2022interleaving}.

One common limitation of these systems is the use of sparse retrievers, such as BM25~\cite{robertson1995okapi}, even though dense retrievers like {Contriever~\cite{izacardunsupervised}} are known to be more effective in general. 
This is largely due to the fact that, unlike sparse retrievers based on keyword matching, dense retrievers rely on query and document embeddings that need to be trained on the target domain~\cite{karpukhin2020dense}. 
% \phseo{???}
% % \dslee{Initial thought for query-document labels was only original question-document, not subquestion-documemt relevance, this is a good point, papers that decompose the question at the start uses the subquestion label, they create an dataset for question decomposition, However, I would like to add that I think question decomposition and step-by-step problem-solving are slightly different approaches. In question decomposition, the question is broken down into smaller, more manageable subquestions, whereas step-by-step problem-solving may not always require explicit decomposition but rather an iterative reasoning process}
% \dslee{In the context of MHQA, however, it is difficult to prepare relevance-labeled query-document pairs needed to train a retriever, as labeling all required documents is resource-intensive and costly.}
% \dslee{In an iterative RAG framework for MHQA, however, labeling multiple distinct documents for initial question is challenging, and preparing documents with relevance labels for retrieval queries is difficult, as reformulated queries can vary across large language models (LLMs), even within the same domain.}
For MHQA, however, it is cost- and labor-intensive to prepare documents labeled with their relevance to respective queries across iterations, because the queries---reformulated questions---can be different for each large language model (LLM) used for answer generation, even for the same domain.

To address this issue, we propose Retriever Supervision with Consistency and Relevance (ReSCORE), a novel method for training a dense retriever for MHQA without labeled documents. 
ReSCORE builds on the intuition that the importance of a document for answering a question is proportional to the probability of an LLM generating both the question and the correct answer given the document. 
In this way, the document's \textit{consistency} with the answer~\cite{izacard2023atlas} and \textit{relevance} to the question are jointly modeled. 
ReSCORE leverages this probability as pseudo-ground truth (\textit{pseudo-GT}) label to train the retriever within an iterative RAG framework.

We demonstrate the efficacy of ReSCORE through experiments on three popular MHQA datasets: MuSiQue \cite{trivedi2022musique}, 2WikiMHQA \cite{ho2020constructing}, and HotpotQA \cite{yang2018hotpotqa}.
The experiments show that a combination of consistency and relevance provides effective supervision for training a dense retriever for MHQA without labeled documents.
The retriever trained using ReSCORE not only improves the retrieval quality but also achieves the SOTA performance on MHQA when integrated into our iterative RAG framework, Iterative Question Answerer with Trained Retriever (IQATR, which is pronounced as ``equator'').
Our key contributions are as follows:
\begin{itemize}[itemsep=2pt, topsep=0pt, parsep=0pt, partopsep=0pt]
    \item We propose ReSCORE, an iterative dense retriever training approach for MHQA without relying on documents labeled with their relevance to respective queries.
    \item We present IQATR, an MHQA system with its retriever trained using ReSCORE. It achieves the SOTA on three popular benchmarks, thereby showcasing the efficacy of ReSCORE.
    \item We provide an in-depth analysis of the effects of various pseudo-GT labels and query reformulation methods.
    % \item We demonstrate the efficacy of ReSCORE through comprehensive analyses across various aspects.
    % \item We present IQATR, a new MHQA system achieving the SOTA performance on three popular MHQA benchmarks.
    % \item We further analyze \dslee{ReSCORE}
\end{itemize}

%% file: figures/01_fig.tex
\begin{figure}[!t]
    \centering
 \includegraphics[width=\linewidth, trim=5.9cm 1.2cm 21.6cm 2.8cm, clip]{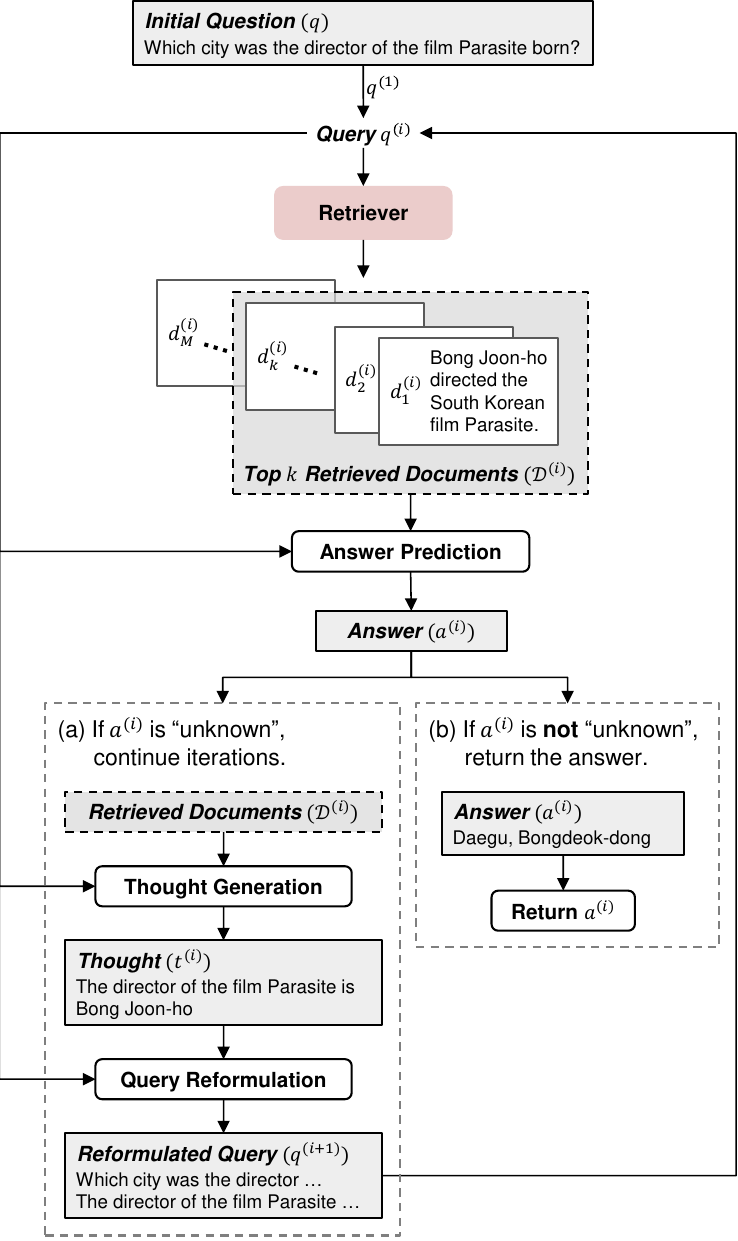}
    \caption{\textbf{Iterative RAG Framework for MHQA.} At iteration $i$, the framework first retrieves top $k$ documents relevant to the current query $q^{(i)}$ to generate an answer $a^{(i)}$. (a) If the answer is "unknown", a thought $t^{(i)}$ is generated as a compact representation of the retrieved documents based on the query $q^{(i)}$. This thought is then used to reformulate the query for the next iteration $q^{(i+1)}$ and continues the next iteration. (b) If $a^{(i)}$ is not "unknown", the iteration ends, and $a^{(i)}$ is returned as the final answer.}
    \vspace{-1em}
    \label{fig:inference}
\end{figure}

%% file: src/02_related_works.tex
\section{Related Work}
% \jpark{End each paragraph with a sentence or two describing the relevance to/distinction from our work.}
% Example
% Optical flow deals with the dense computation of instantaneous motion patterns between two given video frames.
% Starting with the pioneering work of applying neural networks for motion estimation [17, 18], the seminal works such as DCFlow [19], PWC-Net [20] and RAFT [21] introduced the concept of dense correspondence matching between pairs of image patches. 
% Despite their success, the optical flow’s inherent limitations incapable of modeling trajectories and occlusions triggered the recent progress in the point tracking methods.
\textbf{Training Retrievers for RAG} \ \ In the context of RAG, retrieval accuracy plays a critical role in improving the performance of the overall system. 
Several approaches have focused on improving retrieval quality by training retrievers, including supervised training with large labeled datasets~\cite{izacardunsupervised, guu2020retrieval}, and unsupervised training~\cite{izacardunsupervised}.
While these methods primarily concentrate on optimizing a retriever, they often overlook the generation aspect, leading to a domain gap between retrieval and generation tasks.
To bridge this gap, techniques leveraging LLM supervision, LLM-Embedder~\cite{zhang2023language}, Intermediate Distillation~\cite{li2024intermediate}, REPLUG~\cite{shi2023replug} and ATLAS~\cite{izacard2023atlas} have proposed methods that train the retrieval to align with generation, aiming to optimize both processes.
However, these approaches typically focus on single-hop questions and only consider consistency of the document with the answer, overlooking iterative reasoning and MHQA.
In contrast, our approach trains within an iterative framework, emphasizing both the consistency and the relevance of a document, offering a more holistic solution for MHQA.

% Contriever~\cite{izacard2021unsupervised} uses unsupervised data
% %, where the text from the same paragraph serves as both the query and the document
% , but require additional training in unseen domains.
% Intermediate Distillation~\cite{li2024intermediate} prompts the LLM to assign scores to documents, which are then used to train the retriever. 
% Similarly, LLM-Embedder~\cite{zhang2023language} trains the retriever using reward signals from large language models.
% REPLUG~\cite{shi2023replug} and ATLAS~\cite{izacard2023atlas} present a retriever training through LLM probability of predicting the next token. 
% Our approach leverages the LLM probability while incorporating both \textit{relevance} to a query and \textit{consistency} of the answer, addressing a limitation of previous works, which primarily focused on \textit{consistency} alone.} 

% \textbf{Multi-hop Retrieval-Augmented Generation (Multi-hop RAG)}
\noindent\textbf{Iterative RAG} \ \ 
Iterative RAG extends single-hop RAG to tasks requiring multiple reasoning steps across documents~\cite{xiong2020answering}. 
% \phseo{update}
FLARE~\cite{jiang2023active} focuses on adaptively retrieving documents when low-probability tokens are generated. 
To dynamically determine the need for external knowledge, Self-RAG~\cite{asai2023self} trains on a GPT-4~\cite{brown2020language} generated dataset. 
ITER-RETGEN~\cite{shao2023enhancing} incorporates the output from the previous iteration as a retrieval context. 
Another notable method, IRCoT~\cite{trivedi2022interleaving}, extends a Chain of Thoughts iteratively to mimic multi-step reasoning.
Building on IRCoT, Adaptive-RAG~\cite{jeong2024adaptive} improves efficiency by introducing a classifier that dynamically adjusts the number of reasoning steps based on question complexity.
% FLARE~\cite{jiang2023active} adaptively retrieves when low probability tokens are generated 
% Self-RAG~\cite{asai2023self} is trained on a GPT-4~\cite{brown2020language} generated dataset to generate a retrieval token when the external knowledge is required.
% ITER-RETGEN~\cite{shao2023enhancing} uses the previous iteration's model output as context for the next retrieval.
% IRCoT~\cite{trivedi2022interleaving} iteratively retrieves and extends a Chain of Thoughts to imitate a multi-step reasoning process. 
% Adaptive-RAG~\cite{jeong2024adaptive}, building on IRCoT, adds a classifier to determine the complexity of the question, dynamically determining the number of reasoning steps for efficiency.
%only complex questions to a multi-step process for efficiency. 
% \phseo{update}
Adaptive-Note~\cite{wang2024retriever} filters out some of retrieved documents using an LLM to improve precision.
% \dslee{
While the aforementioned works excel in iterative RAG, none of them focus on training retrievers, which is a crucial element and rely either on traditional sparse retrievers or a dense retriever pretrained on a different dataset.
% }
% However, none of these works directly train the retriever for MHQA. 
% Consequently, they rely on traditional sparse retrievers, such as BM25, or a pretrained dense retriever like Contriever~\cite{izacard2021unsupervised},
%, a dense retriever not fine-tuned for MHQA,
% which is suboptimal due to domain shifts.
%semantic understanding. 
% \dslee{
In contrast, we train a dense retriever directly within the iterative RAG system, and allow the retriever to effectively adapt to the target domain.
% enabling it to better handle the domain-specific challenges of MHQA and improve retrieval quality across multiple reasoning steps.
% }
% The studies most closely related to ours are MDR~\cite{xiong2020answering} and LOUVRE~\cite{seonwoo2021weakly}, which also employ iterative RAG while training a retriever for MHQA.
% However, both approaches require negative document labels and focus only on 2-hop questions. 
% In contrast, we aim to perform MHQA with more than 2-hops and train without the need for document labels.

% \dslee{
% Iterative RAG extends single-hop RAG to tasks requiring multiple reasoning steps across documents~\cite{xiong2020answering}. 
% Recent works, such as FLARE~\cite{jiang2023active} and Self-RAG~\cite{asai2023self}, introduce adaptive retrieval mechanisms, but Flare can cause overhead in ambiguous contexts, and Self-RAG's reliance on self-reflection doesn't directly improve retriever accuracy.
% Iterative retrieval-generation frameworks like IRCoT~\cite{trivedi2022interleaving}, ITER-RETGEN~\cite{shao2023enhancing}, Adaptive-Note~\cite{wang2024retriever}, and Adaptive-RAG~\cite{jeong2024adaptive} alternate between retrieval and generation to refine context. 
% However, their reliance on sparse retrievers (e.g., BM25) or Contriever~\cite{izacard2021unsupervised} not fine-tuned for MHQA, limits their ability to capture semantic relationships, leading to reduced performance in MHQA. 
% We enhance retriever accuracy in iterative RAG framework to improve performance in MHQA.
% }

\input{figures/02_fig}
\noindent\textbf{Training with LLM Supervision} \ \  In recent years, training smaller models with LLM supervision has become a common and effective approach, especially when human annotation is limited or unavailable. 
One notable example is CoT-Distill~\cite{shridhar2022distilling}, which utilize teacher model generated Chain-of-Thought dataset to train smaller models. 
In a similar vein, Self-RAG~\cite{asai2023self} employs a dataset curated by GPT-4~\cite{brown2020language} to learn a classifier deciding when to retrieve.
Moreover, Intermediate Distillation~\cite{li2024intermediate}, Promptagator~\cite{dai2022promptagator}, and RankVicuna~\cite{pradeep2023rankvicuna} explore the use of teacher model generated document ranking lists to guide the training process.
Other works, such as DistilBERT~\cite{sanh2019distilbert}, which is a smaller version of BERT trained by leveraging the hidden states vector of a teacher model. 
Similarly, ATLAS~\cite{izacard2023atlas} uses token probabilities from the teacher model to train a retriever.
% \phseo{update}
% \dslee{
To the best of our knowledge, this study is the first to leverage an LLM for training a retriever within an iterative RAG framework for MHQA.

%% file: figures/02_fig.tex
\begin{figure*}[!ht]
    \centering
    \includegraphics[width=1.0\textwidth, trim=2.2cm 7.9cm 9.3cm 4.3cm, clip]{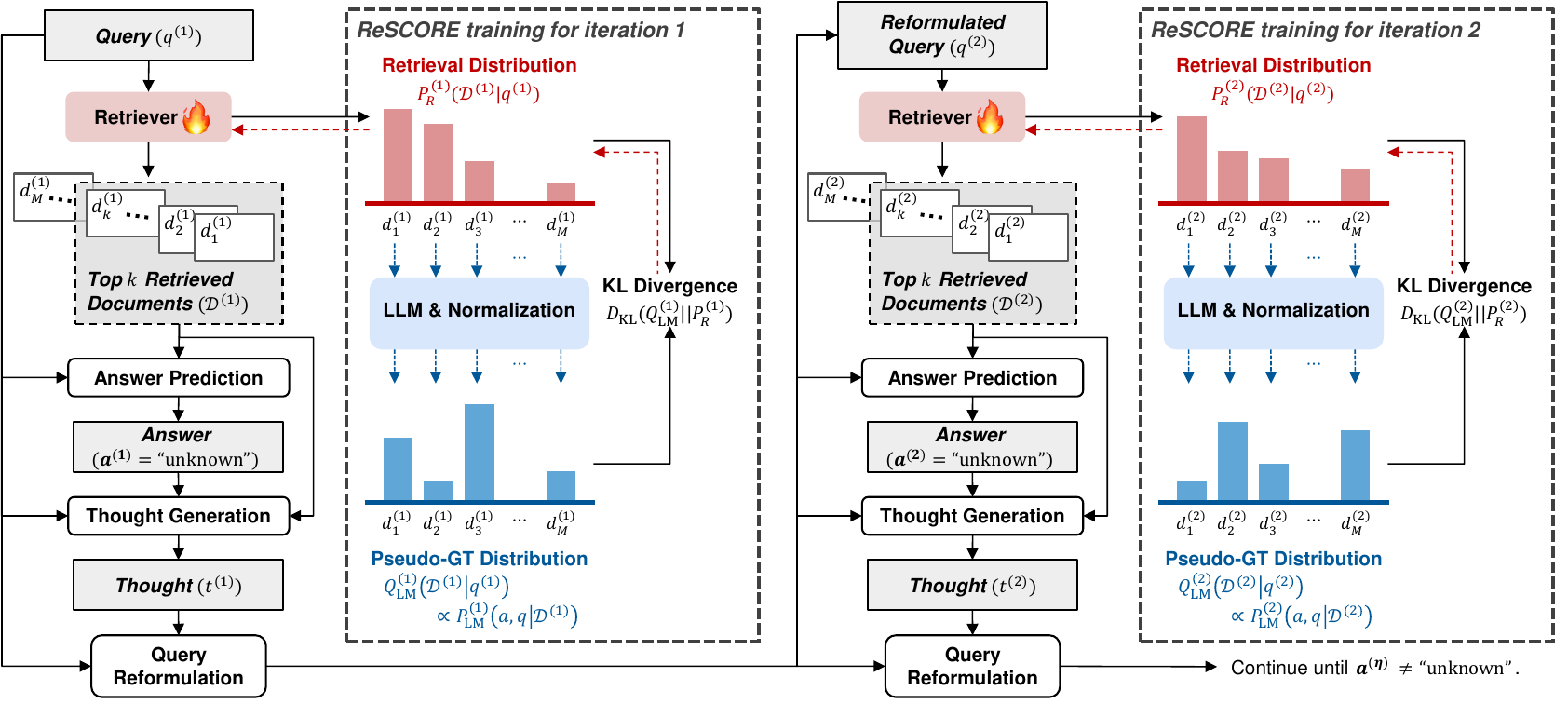}
    \caption{\textbf{Overview of ReSCORE.} 
    At each iteration $i$ within a iterative RAG process, the retriever receives gradients from the KL-Divergence loss of the retrieval distribution $P_R^{(i)}$ against the pseudo-GT distribution $Q_\text{LM}^{(i)}$, which is derived from the LLM probabilities of question and answer given each document $d_j^{(i)}$ with normalization.
    The number of iterations is dynamically determined by the LLM and the process ends if the LLM predicts an answer which is not ``unknown''.
    The red dashed lines represents gradient flows for the retriever.
    }
    \vspace{-0.85em}
    \label{fig:training}
\end{figure*}

%% file: src/03_methods.tex
\section{Methods}
\subsection{Iterative RAG Framework}
\label{sec:inference}

Given a question $q$, the goal of MHQA is to generate the answer $a$ leveraging knowledge from a document database $\mathcal{D}$ from which relevant documents are retrieved.
Notably, the question $q$ can be answered only if the complete set of relevant documents $\mathcal{D}^*=\{d^*_1, \dots,d^*_h\} \subseteq \mathcal{D}$ is accurately identified and utilized.
To tackle this problem, we adopt an iterative reasoning process, following previous studies~\cite{trivedi2022interleaving,jeong2024adaptive}, where the system iteratively retrieves relevant documents and refines the answer based on the retrieved information as illustrated in Fig.~\ref{fig:inference}.

Specifically, given the question $q$, which we designate as the first query ${q}^{(1)}$, we retrieve a set of $k$ documents $\mathcal{D}^{(1)}=\{d_1^{(1)}, \dots,d_k^{(1)}\} \subseteq \mathcal{D}$.
$\mathcal{D}^{(1)}$ is then incorporated into a predefined prompt for the LLM. 
% The prompt instructs the LLM to either output an answer based on the information in $\mathcal{D}^{(1)}$, thereby terminating the iterative reasoning process, or defer answer generation to retrieve additional documents, as illustrated by Fig.~\ref{fig:inference}(a) and (b), respectively.
The prompt instructs the LLM to either defer answer generation to retrieve additional information, or predict an answer ${a}^{(1)}$ based on the sufficiency of the information in $\mathcal{D}^{(1)}$, thereby terminating the question-answering process, as illustrated by (a) and (b) in Fig.~\ref{fig:inference}, respectively.
If the LLM decides to retrieve additional documents by predicting ``unknown'' as the answer, the system constructs a compressed representation of the retrieved documents $\mathcal{D}^{(1)}$, referred to as a \textit{thought} $t^{(1)}$.
To achieve this, we prompt the LLM to construct a single sentence distilling the key information required to answer the initial question $q$ from the retrieved documents $\mathcal{D}^{(1)}$.
This technique, adopted from \cite{trivedi2022interleaving}, allows us to maintain the retrieved information in a compact form, which is then utilized during subsequent iterations in answer generation.
Finally, the system reformulates the query $q^{(1)}$ into a new query $q^{(2)}$ highlighting unresolved aspects of $q^{(1)}$ in $\mathcal{D}^{(1)}$ requiring additional information.
This reformulated query $q^{(2)}$ then guides the retrieval of additional documents in the next iteration.

% These documents $\mathcal{D}^{(0)}$ are then incorporated into a predefined prompt for the LLM. 
% The prompt determines whether the LLM should generate an answer\dslee{prediction} based on the sufficiency of the information in \( \mathcal{D}^{(0)} \) Fig.~\ref{fig:inference}(a), or if it should provide a thought \( t^{(1)} \) Fig.~\ref{fig:inference}(b).
% IRCoT previous works used the term thought, with citation, document dependent summarization 
% The thought \( t^{(1)} \) is formulated as a single sentence that distills the key points from the retrieved documents, and helps answering the initial question\cite{trivedi2022interleaving}.
% At the final phase \dslee{of this hop}, we rewrite the question \( q^{(0)} \) into a new question \( q^{(1)} \). 
% This new question \( q^{(1)} \) highlights aspects of the original query that remain unresolved or require more specific information, and it guides the retrieval of additional relevant documents in the next iteration.

In the next iteration, the refined query $q^{(2)}$ is used to retrieve a new set of $k$ documents $\mathcal{D}^{(2)}=\{d_1^{(2)}, \dots,d_k^{(2)}\} \subseteq \mathcal{D}-\mathcal{D}^{(1)}$.
These retrieved documents are then provided to the LLM along with the thought $t^{(1)}$, which either outputs a final answer or continues the iterative process by generating a new thought $t^{(2)}$ and a further reformulated query $q^{(3)}$.
% to generate the answer. The iterative process continues, generating a new thought $t^{(1)}$ and a further refined question $q^{(2)}$ if the LLM determines that additional information is needed. 
More generally, at each iteration $i$, a set of $k$ new relevant documents $\mathcal{D}^{(i)}$ is retrieved based on the query $q^{(i)}$.
Then, the LLM either generates the final answer based on the retrieved documents $\mathcal{D}^{i}$, as well as all available thoughts $t^{(1)},\dots,t^{(i-1)}$ or continues the process with a new thought $t^{(i)}$ and a reformulated query $q^{(i+1)}$.
\subsection{Training Retriever for Iterative RAG}
% \subsection{Answer-guided Retriever Training with Iteratively Supporting Evidence (\dslee{ReSCORE})}
\label{sec:training}

A key component of this iterative RAG framework is the retriever, which must ensure the retrieval of documents that provide relevant and complementary information across iterations to support multi-hop reasoning.
However, collecting labeled documents for retriever training is labor- and cost-intensive. 
To address this limitation, we propose ReSCORE, a novel framework for retriever training without document labels. 
In ReSCORE, a retriever is trained for MHQA using pseudo-GT labels generated by leveraging an LLM, as illustrated in Fig.~\ref{fig:training}.

\noindent\textbf{Generating Pseudo-GT Labels} \ \ 
As labels for relevant documents are unavailable, it is essential to devise a method to identify which documents are required to the input question to effectively train the retriever. 
Specifically, we measure the distribution $Q_\text{LM}^{(i)}(d_j^{(i)}\mid q^{(i)})$, which represents the likelihood of retrieving a document $d_j^{(i)}$ given a query $q^{(i)}$ at iteration $i$.
To achieve this, we leverage an LLM inspired from \cite{izacard2023atlas} capturing the intuition that $Q_\text{LM}^{(i)}(d_j^{(i)}\mid q^{(i)})$ for a document $d_j^{(i)}$ is proportional to the probability that the LLM generates both the question $q$ and the corresponding answer $a$ given $d_j^{(i)}$. 
Formally, this is expressed as:
\begin{align}
    Q_{\text{LM}}^{(i)}(d_{j}^{(i)} \mid q) &\propto P_{\text{LM}}^{(i)}(a, q \mid d_{j}^{(i)})  \label{eq:q_lm} \\
    &= P_{\text{LM}}^{(i)}(q \mid d_{j}^{(i)}) \cdot P_{\text{LM}}^{(i)}(a \mid q, d_{j}^{(i)}) \label{eq:decomp}
\end{align}
where $P_\text{LM}$ denotes the probability of a token sequence as computed by the LLM.

The advantage of our approach lies in its ability to evaluate not only the \textit{relevance} of the document to the question but also its \textit{consistency} in answering the question. 
The probability in Eq.~\eqref{eq:q_lm} can be further decomposed into two components using the chain rule as in Eq.~\eqref{eq:decomp}.
The former represents the probability of generating the question from the document, capturing the relevance.
The latter represents the probability of predicting the correct answer to the question with the document, assessing the consistency.
While $P_{\text{LM}}^{(i)}(a \mid q, d_{j}^{(i)})$ appears more directly aligned with the QA training objective, this term alone often fails to capture the relevance of the document to the question. 
Notably, determining whether a document is consistent for answering a given question is often challenging, even for humans. 
This can lead to high LM scores, even when there are only superficial word-level alignments between the document and the answer, which may not necessarily reflect true relevance.
For instance, $P_{\text{LM}}^{(i)}(a \mid q, d_{j}^{(i)})$ for a document titled “2006 FIFA World Cup” is higher than that for the GT documents for the question:
“In what year did the studio behind Toy Story release its first feature film after being acquired by Disney?”
This occurs because, while the document is irrelevant, it contains the token “2006”, which is the correct answer\footnote{Pixar—the studio behind Toy Story—was acquired by Disney in January 2006, and its first post-acquisition film, Cars, was released in May 2006.}.
% $P_{\text{LM}}^{(i)}(a, q \mid d_{j}^{(i)})$ 
% \dslee{
In contrast, 
Eq.~\eqref{eq:decomp} 
additionally incorporates 
$P_{\text{LM}}^{(i)}(q \mid d_{j}^{(i)})$,
which is low for topically unrelated documents, thereby down-weighting them in the in the final scores.
% }
% which tends to be low for documents that are topically unrelated to the question, thereby reducing their influence in the scoring.
% , which is low when the document is topically unrelated and unlikely to predict the question tokens, thereby filtering it out.
% \jpark{
% For instance, $P_{\text{LM}}^{(i)}(a \mid q, d_{j}^{(i)})$ for a document “2006 FIFA World Cup”
% is higher than that for the GT documents for the question: 
% “In what year did the studio behind Toy Story release its first feature film after being acquired by Disney?”
% While the document is irrelevant, it contains the token “2006”, which is the correct answer to the question. 
% In contrast, $P_{\text{LM}}^{(i)}(a, q \mid d_{j}^{(i)})$ additionally incorporates $P_{\text{LM}}^{(i)}(q \mid d_{j}^{(i)})$, which evaluates the document's relevance to the question.
% }
Furthermore, a document's relevance to a question itself does not imply that it provides adequate information for answering the question, as it overlooks the consistency to the answer.
By explicitly modeling both consistency and relevance, our method trains a retriever to retrieve the documents necessary for answering a given question.

\noindent\textbf{Training Loss Function} 
Given the distribution $Q_\text{LM}$ as the ground truth, we train the retriever by minimizing the Kullback-Leibler (KL) divergence over all QA pairs $(q_n,a_n)$ and iterations $i$:
\begin{equation*}
% \mathcal{L} = 
\sum_{n=1}^{N}\sum_{i=0}^{\eta_n}D_{\text{KL}} \left( Q_{\text{LM}}^{(i)}(\mathcal{{D}}^{(i)} \mid q_n^{(i)}) \parallel P_R^{(i)}(\mathcal{{D}}^{(i)} \mid q_n^{(i)}) \right),
\end{equation*}
% \needspace{5\baselineskip}
where $N$ is the number of QA pairs in the training set, $\eta_n$ is the number of iterations determined by the LLM for each question $q_n$,
% \jpark{$q_r$?} 
and $P_R^{(i)}$ is the document distribution for retrieval at iteration $i$. 
The distribution $P_R^{(i)}$ is computed by applying the Softmax function on the dot products between the question vector and each document vector in the database $\mathcal{D}$, \textit{i.e.},
\vspace*{-0.2em}
\begin{equation*}
P_R^{(i)}(d_j^{(i)} \mid q_n^{(i)}) = \text{Softmax} \left( \mathbf{d}_j^{(i)} \cdot \mathbf{q}_n^{(i)} \right)
\end{equation*}
where $\mathbf{d}_j ^{(i)}=\text{Embed}_{\text{doc}}(d_j^{(i)})$ is a document embedding and $\mathbf{q}_n^{(i)}=\text{Embed}_{\text{query}}(q_n^{(i)})$ is a query embedding.

Note, the GT answer $a_n$ for each instance is used to compute $Q_{\text{LM}}^{(i)}$, which serves as the distribution the retriever aims to learn.
However, calculating the distribution $Q_{\text{LM}}^{(i)}(\mathcal{{D}}^{(i)} \mid q_n^{(i)})$ over the entire database $\mathcal{D}$ is computationally prohibitive due to the large size of $\mathcal{D}$ and the high computational cost of the LLM. 
Thus, at each iteration $i$, we sample the top $M\ll|\mathcal{D}|$ documents based on the retriever scores and compute $Q_\text{LM}^{(i)}$ only on these sampled documents.

%% file: src/04_experiments.tex
\section{Experiment}
\subsection{Settings}
\textbf{Datasets} \ \ 
We conduct our experiments on three popular MHQA datasets: MuSiQue \cite{trivedi2022musique}, 2WikiMHQA \cite{ho2020constructing}, and HotpotQA \cite{yang2018hotpotqa}. 
Each dataset contains complex question structures that require reasoning across multiple documents, making them ideal for evaluating multi-hop retrieval and question-answering capabilities. 
Following prior works~\cite{trivedi2022interleaving, jeong2024adaptive, 8733051}, experiments are conducted on subsampled versions of the validation and test sets, as well as the retrieval database. 
% , which is constructed in accordance with the setup used in these previous studies~\cite{trivedi2022interleaving, jeong2024adaptive}. 
These datasets come with GT document labels, which are not used for training our model.
% \phseo{they come with GT labels but we do not use?}

\noindent\textbf{Models} \ \ 
We take as baselines the best existing models for MHQA: ReAcT~\cite{yao2022react}, FLARE~\cite{jiang2023active}, Self-RAG~\cite{asai2023self}, Adaptive-Note~\cite{wang2024retriever}, IRCoT~\cite{trivedi2022interleaving}, and Adaptive-RAG~\cite{jeong2024adaptive}.
We then establish our own baseline models by implementing the iterative RAG framework described in Sec.~\ref{sec:inference}, integrating Llama-3.1-8B-Instruct~\cite{touvron2023llama} with BM25~\cite{robertson1995okapi} and Contriever~\cite{izacardunsupervised}
trained on the MS-MARCO dataset~\cite{bajaj2018msmarcohumangenerated}.\footnote{Unlike existing works, we employ Llama to address Flan-T5's slow inference and GPT-3.5's cost issue.
% difficulty in scoring with probability due to long few-shot prompts.
Also, Contriever is one of the best-performing dense retrievers. It is typically trained on MS MARCO and fine-tuned on the target domain.}
% We then demonstrate the effectiveness of our approach by fine-tuning the retriever of this baseline model.
Lastly, we prepare our model---Iterative Question Answerer with Trained Retriever (IQATR)---by fine-tuning Contriever in our baseline model using ReSCORE.

\noindent\textbf{Evaluation Metrics} \ \ 
To assess the QA performance of our approach, we adopt two standard metrics for MHQA: Exact Match (EM) and F1 score.
These metrics are applied at the answer level, using the official evaluation protocol provided in each dataset.
% In addition, we directly evaluate the retriever based on recall at $k$ (R@$k$), which measures the proportion of GT relevant documents successfully retrieved within the top $k$ candidates.
% While R@$k$ measures the retrieval quality, it fails to capture how complementary the retrieval results from different iterations are for answering multi-hop questions.
% Therefore
% \dslee{Specifically}, we also propose to measure multi-hop recall at $k$ at iteration $i$ ($\text{MHR}_i\text{@}k$), which is 
% \dslee{
To assess the retrieval performance within our iterative RAG framework, 
% we note that the standard recall metric does not capture the retrieval performance at each iteration in iterative RAG.
% Therefore, 
we introduce a metric called multi-hop recall at $k$ (MHR@$k$), measuring recall across iterations. 
Specifically, we compute the MHR@$k$ for iteration $i$, denoted as $\text{MHR}_i\text{@}k$, by
% }
\vspace*{-1em}
\begin{equation}   \text{MHR}_{i}\text{@}k = \frac{\left| \mathcal{D}^* \cap \bigcup_{l=1}^i \mathcal{{D}}^{(l)} \right|}{\left| \mathcal{D}^* \right|}
\end{equation}
where $\mathcal{D}^*$ is the set of GT supporting documents, and $\bigcup_{l=1}^i \mathcal{{D}}^{(l)}$ is the union of retrieved documents up to iteration $i$.
%with $\mathcal{\hat{D}}^{(l)}$ being the set of $k$ retrieved documents at iteration $l$.
This measures the cumulative recall at iteration $i$ as the ratio of GT supporting documents retrieved up to iteration $i$ to the total number of the GT supporting documents.
% For multi-hop recall (MHR), the cumulative recall at iteration \( i \) is defined as the ratio of the relevant documents retrieved up to and including iteration \( i \) to the total number of relevant documents in the entire query set.
% where \( \mathcal{D}_{\text{relevant}, i} \) denotes the set of relevant documents that have not yet been retrieved at hop \( i \), excluding those retrieved in previous hops. \( \mathcal{\hat{D}}_i \) represents the subset of the documents retrieved at hop \( i \).
\input{tables/01_sota_table}

\noindent\textbf{Implementation Details} \ \
%\jpark{Move to Sec 4}
We train the question embedder while keeping the document embedder frozen throughout the process. 
To compute the document distribution, we format the question, answer, and document into a predefined prompt, as described in Section~\ref{sec:appendix}. 
For loss calculation, we use the top $M$ = 32 documents, while for inference, we select the top $k$ = 8 documents. 
The maximum number of iterations $\eta_n$ is set to 6, and the batch size to 16.
Temperature scaling is applied to control the output distributions of the LLM, with a temperature value of 0.1, which is selected among 1, 0.1, and 0.01. 
% Among these, a temperature of 0.1 produces the most stable results.
We use the AdamW optimizer and two NVIDIA A100 GPUs (40GB memory). 
The initial learning rate is set to $1 \times 10^{-6}$ and is exponentially decayed at every 100 iterations by a factor of 0.9. 
The training continues until the validation loss stops improving within an epoch.
Additionally, in accordance with the MHQA requirements, which involve reasoning over at least two hops, we set a minimum iteration limit of 2, in both training and inference of IQATR, inspired by Adaptive-RAG~\cite{jeong2024adaptive}.

\subsection{Results and Analysis}
\subsubsection{Efficacy of ReSCORE}
% We compare our models incorporated with a fine-tuned retriever with the baseline models and the existing state-of-the-art (SOTA) methods in Tab.~\ref{tab:sota}.
We first compare IQATR, equipped with a retriever fine-tuned by ReSCORE, against baseline models and existing SOTA methods in Tab.~\ref{tab:sota}.
% The results first present that our baseline model performs better with BM25 retriever than with a pretrained Contriever, which is consistent with the findings in the previous studies~\cite{}. 
The results first present that baseline models perform 
better with the sparse BM25 retriever than with a pretrained Contriever.
% \dslee{
This can be attributed to the fact that Contriever was not trained on domain-specific data~\cite{izacardunsupervised}.

% }
% \dslee{in our case MHQA, due to the lack of task-specific fine-tuning, as highlighted in the Contriever paper, where it is noted that Contriever fails in scenarios with unseen data, such as COVID-related tasks~\cite{izacard2021unsupervised}.}
Although BM25 performs better initially, however, its training-free nature limits its potential for further improvement. 
In contrast, the document representations of Contriever can be enhanced through fine-tuning, enabling greater adaptability and performance gains. 
Consequently, when fine-tuned with ReSCORE, the model demonstrates significant improvements across all metrics on all three benchmarks, achieving SOTA performance.
% In addition, we enforce a minimum of two iterations per question, consistent with 
% MHQA tasks
% ~\cite{jeong2024adaptive}, \dslee{which adapts multi-step for complex questions}, during both training and evaluation, since MHQA tasks typically involve at least two reasoning hops.
% This approach leads to further improvements, achieving state-of-the-art performance across all three MHQA benchmarks.

In addition, we test the proposed method, ReSCORE, with other existing iterative MHQA methods, including Self-RAG~\cite{asai2023self}, FLARE~\cite{jiang2023active}, and Adaptive-Note~\cite{wang2024retriever}. 
These frameworks are re-implemented using Llama and Contriever to avoid costs for API calls. 
Tab.~\ref{tab:base_model_comparison} presents the MHQA performance in terms of EM and F1, as well as retrieval performance measured by $\mathrm{MHR}_i\text{@}k$ with $k=8$ and varying $i$.
Note that $\eta_n$ represents the total number of iterations, which varies for each question.
% \dslee{Note that $\eta_n$ represents the total number of iterations, which varies for each question based on the number of iterations performed}
The results clearly demonstrate that ReSCORE consistently enhances both MHQA and retrieval performances across all methods and benchmarks, highlighting its broad applicability.
Notably, the improvements in $\mathrm{MHR}_i\text{@}8$ become bigger as $i$ increases.
The $\mathrm{MHR}_i\text{@}8$ scores in the baseline models are bounded even though $i$ increases whereas the scores with the retrievers fine-tuned with ReSCORE continue to improve as $i$ grows.
This signifies that ReSCORE effectively trains the retriever to identify documents that complement those already retrieved.

\input{tables/03_comparison_table}
\input{tables/02_relevance_table}
\subsubsection{Analysis of Pseudo-GT Labels}
\label{sec:relevance_scores}
% \dslee{TODO: maybe add relative average gain specifically, showing how each metric performs}
% \dslee{
We next demonstrate the effectiveness of using the proposed pseudo-GT labels for fine-tuning the retriever
% We investigate how pseudo-GT labels contribute to the performance, 
%Given the pretrained baseline retriever,
% }
by comparing the results of three LLM-based re-ranking methods, including the proposed approach: 
$P_{\text{LM}}(q \mid d_j)$, 
$P_{\text{LM}}(a \mid q, d_j)$ and $P_{\text{LM}}(q, a \mid d_j)$.
The first question probability, $P_{\text{LM}}(q\mid d_j)$, evaluates the relevance of the document $d_j$ to the question $q$. 
The second answer probability, $P_{\text{LM}}(a\mid q, d_j)$, measures the consistency of the document in answering the question.
Finally, the third approach, $P_{\text{LM}}(q,a\mid d_j)$, which is adopted as the pseudo-GT labels in ReSCORE, jointly considers both relevance and consistency, providing a comprehensive metric for training a retriever. 
For this experiment, we simply measure the standard recall on re-ranked results in a single iteration.
% Since this is not an iterative retrieval setting, we use standard recall as the evaluation metric.

% 5.37 -23.8 14.4

The results in Tab.~\ref{tab:document_relevance} demonstrate that re-ranking documents using the question probability improves recalls across all three datasets by an average of $5.37\%$.
This highlights the critical role of considering document relevance to the question in retrieval for MHQA.
Interestingly, however, re-ranking documents solely based on the answer probability significantly degrades $23.8\%$ from the baseline performance on average.
This decline is primarily due to an increase in false positives, where irrelevant documents are erroneously assigned high consistency scores because of their superficial alignment with the answer confusing the LLM.

Finally, we tested our proposed approach, which uses the QA probability, combining relevance and consistency. 
Note that this QA probability can be factorized as the product of the question and answer probabilities, $P_{\text{LM}}(q \mid d_j)\cdot P_{\text{LM}}(a \mid q, d_j)$.
The results show approximately $14.4\%$ improvements on average across the benchmarks compared to the baseline.
While the answer probability by itself seemed ineffective, its combination with the question probability becomes powerful, as it evaluates the consistency among relevant documents, with irrelevant ones already filtered out due to their low question probabilities.

\input{tables/04_training_signal_table}

\input{figures/03_fig}

\subsubsection{Pseudo-GT vs. GT Labels}
To further evaluate the quality of pseudo-GT labels in ReSCORE, we fine-tune retrievers with GT labels.
% For this fine-tuning, a retriever is trained to assign high scores to all labeled GT documents at the initial iteration in a single step.
% \dslee{
For this fine-tuning, a retriever is trained in a single step, treating all labeled GT documents as positives. Here,  InfoNCE loss function from DPR~\cite{karpukhin2020dense} and Contriever~\cite{izacardunsupervised} is employed in line with the common practice for fine-tuning dense retrievers.
% \phseo{is there a name for the loss function?}
% }
While it might be hypothesized that such models serve as an upper bound for ReSCORE, experimental results in Tab.~\ref{tab:multistep_ablation} reveal that ReSCORE-trained models outperform these models, achieving superior results in both MHQA and multi-hop retrieval metrics.
This occurs because the model trained with GT labels forces the query to align with multiple documents simultaneously.
Note that, the GT document distribution remains fixed. 
In this context, a contrastive loss with treating all GT documents as positive attempts to align the query with potentially distant multiple GT documents simultaneously. 
%\dslee{
For example, consider the query: 
% \textit{
"Who is the first president of the country where Billie Eilish’s favorite food originates from?"
% } 
To answer this, multiple documents each containing information on 
% \textit{
"Billie Eilish,"
% }
% \textit{
"Avocado,"
% }
and 
% \textit{
"Presidents of Mexico"
% }
are required. 
When training with GT documents, the query encoder would attempt to align the query embedding with the centroid of these distant document embeddings, which may not be effective for retrieving any of the documents. 
% There might be an appropriate loss function to mitigate such an issue. 
% However, designing such a loss function is non-trivial, as the centroid of multiple GT documents represents the optimal solution under a fixed document distribution for single-step retrieval. 
While GT labels enhance initial retrieval results, they show limited effectiveness in the iterative process, as evidenced by the bounded MHR scores for $i\ge2$. 
In contrast, our method employs an iterative retrieval process, enabling the progressive retrieval of distant GT documents across multiple steps. 
This iterative approach inherently addresses the limitations of single-step retrieval by gradually complementing the retrieval results as $i$ increases.
%}
% Note that, while these GT documents may partially share a concept, they necessarily contain distinct information to fulfill the complementarity required in MHQA.
% This leads to potentially distant document embeddings, resulting in suboptimal performance.
% Additionally, while GT labels enhance initial retrieval results, they show limited effectiveness in the iterative process, as evidenced by the bounded MHR scores for $i\ge2$. 
% In contrast, the retriever trained with ReSCORE achieves consistent gains in MHR as $i$ increases.

\input{tables/05_rewriting_table}

This is further illustrated in Fig.~\ref{fig:graph}, which depicts the proportion of questions for which all relevant documents are successfully retrieved.
As observed, retrievers trained with GT annotations achieve higher rates in the initial iteration (blue lines) because the training procedure pushes the question embedding towards all relevant documents simultaneously. 
However, ReSCORE-trained retrievers quickly surpass these rates as $i$ increases, achieving significantly higher rates of retrieving all relevant documents (red lines) thanks to the incorporation of the iterative process within ReSCORE.

\subsubsection{Ablations on Query Reformulation}
% \phseo{
We perform an ablation study to evaluate the effectiveness of various query reformulation methods.
The first method, None, uses the original question $q$ as the query at every iteration without any reformulation, serving as a lower bound. Another method, LLM-rewrite, prompts an LLM to rewrite the query $q^{(i)}$ into a refined query $q^{(i+1)}$, focusing on unresolved aspects based on the current retrieved documents $\mathcal{D}^{(i)}$. Finally, Thought-concat appends the current thought $t^{(i)}$ to the query, constructing the updated query as $q^{(i+1)}=[t^{(i)};q^{(i)}]$, where 
$[a;b]$ denotes the concatenation of $a$ and $b$.
% }

% \phseo{
The results in Tab.~\ref{tab:rewriting} show that both query reformulation methods improve retrieval and MHQA performance. 
Thought-concat achieves larger gains on MuSiQue and HotpotQA, while LLM-rewrite performs slightly better on 2WikiMHQA. 
This difference stems from question complexity: LLM-rewrite works well for simpler queries (\textit{e.g.}, 2WikiMHQA with 11.7 tokens on average) but struggles with complex ones (\textit{e.g.}, MuSiQue and HotpotQA with 17.9 and 16.0 tokens, respectively), often losing focus. 
In contrast, Thought-concat benefits from LLMs' strength in summarization and allows error recovery in subsequent iterations, as the original question remains as a part of the reformulated query.

%% file: tables/01_sota_table.tex
\begin{table*}[t]
\centering
\small
\begin{tabular}{lcccccc}
\toprule
 & \multicolumn{2}{c}{\textbf{MuSiQue}} & \multicolumn{2}{c}{\textbf{HotpotQA}} & \multicolumn{2}{c}{\textbf{2WikiMHQA}} \\
\textbf{Model} & \textbf{EM} & \textbf{F1} & \textbf{EM} & \textbf{F1} & \textbf{EM} & \textbf{F1} \\
\midrule
ReAcT (GPT-3.5+BM25)$\dagger$ & 10.2 & 19.7 & 36.0 & 46.9 & 28.0 & 37.3 \\ 
FLARE (GPT-3.5+BM25)$\dagger$ & 11.2 & 18.7 & 36.4 & 47.8 & 31.8 & 42.8 \\ 
Self-RAG (GPT-3.5+BM25)$\dagger$ & 10.6 & 19.2 & 33.8 & 44.4 & 24.4 & 30.8 \\ 
Adaptive-Note (GPT-3.5+BM25)$\dagger$ & 13.2 & 24.2 & \underline{45.6} & \underline{58.4} & 43.2 & 54.2 \\ 
IRCoT (Flan-T5-XL+BM25)$\ddagger$ & 22.0 & \underline{31.8} & 44.4 & 56.2 & \underline{49.7} & \underline{54.9} \\ 
Adaptive-RAG (Flan-T5-XL+BM25)$\ddagger\ddagger$ & \textbf{23.6} & \underline{31.8} & 42.0 & 53.8 & 40.6 & 49.8  \\ 
\noalign{\vskip 1.5pt}
\cdashline{1-7}[2pt/1pt]
\noalign{\vskip 2pt}
Our Baseline (Llama-3.1-8B+BM25) & 15.2 & 23.6 & 42.2 & 55.7 & 44.6 & 52.2 \\ 
 Our Baseline (Llama-3.1-8B+Contriever) 
& 15.2 & 23.8 & 39.4 & 52.3 & 32.8 & 41.6 \\ 
IQATR (Llama-3.1-8B+Contriever trained w/ ReSCORE)
& \underline{23.4} & \textbf{32.7} & \textbf{47.2} & \textbf{59.3} & \textbf{50.0} & \textbf{59.7} \\
\bottomrule
\end{tabular}
\caption{\textbf{
Comparisons to State-of-the-Art Iterative RAG Frameworks on three MHQA benchmarks.}
EM and F1 scores are measured on each dataset.
$\dagger$ Scores are sourced from~\cite{wang2024retriever}.
$\ddagger$ Scores are reproduced using the official codes.
$\ddagger\ddagger$ Scores are sourced from the original paper~\cite{jeong2024adaptive}. 
We conducted significance tests at 
$p=0.05$, confirming IQATR's superiority (details in \cref{sec:stats}).
}
\label{tab:sota}
\end{table*}

%% file: tables/03_comparison_table.tex
\begin{table}[t!]
\centering
\small
\begin{tabular}{lccccc}
\toprule
 & \multicolumn{2}{c}{\textbf{QA}} & \multicolumn{3}{c}{\textbf{$\text{MHR}_{i}@\text{K}$}} \\
\textbf{Model} & \textbf{EM} & \textbf{F1} & $i=1$ & $i=2$ & $i=\eta_n$  \\
\midrule 
\multicolumn{6}{c}{\textbf{MuSiQue}} \\
\midrule
$\text{Self-RAG}^{*}$   & 1.2 & 8.2 & 25.8 & 25.8 & 25.8 \\ 
\ \ +ReSCORE & 2.8 & 10.8 & 24.9 & 31.6 & 31.6 \\ 
\cdashline{1-6}[2pt/1pt]
\noalign{\vskip 2pt}
FLARE & 7.3 & 13.3 & 31.0 & 37.1 & 37.1 \\ 
\ \ +ReSCORE  & 8.2 & 15.3 & 30.9 & 40.1 & 43.3 \\
\cdashline{1-6}[2pt/1pt]
\noalign{\vskip 2pt}
Adaptive-Note  & 9.6  & 17.7 & 44.9 & 50.2 & 50.2 \\
\ \ +ReSCORE  & 11.2 & 20.5 & 45.1 & 49.8 & 55.3 \\ 
\cdashline{1-6}[2pt/1pt]
\noalign{\vskip 2pt}
Our Baseline & 15.2 & 23.8 & 44.9 & 51.6 & 51.6 \\
\ \ +ReSCORE & 23.4 & 32.7 & 46.8 & 63.0 & 65.2 \\ 
\midrule
\multicolumn{6}{c}{\textbf{HotpotQA}} \\
\midrule
$\text{Self-RAG}^{*}$ & 5.6 & 17.9 & 36.1 & 36.5 & 36.5 \\ 
\ \ +ReSCORE  & 8.7 & 19.2 & 33.8 & 37.2 & 37.2 \\ 
\cdashline{1-6}[2pt/1pt]
\noalign{\vskip 2pt}
FLARE   & 27.5 & 38.9 & 37.2 & 48.4 & 48.4 \\
\ \ +ReSCORE  & 31.4 & 42.5 & 39.2 & 48.5 & 51.7 \\ 
\cdashline{1-6}[2pt/1pt]
\noalign{\vskip 2pt}
Adaptive-Note  & 42.0 & 55.3 & 44.8 & 49.8 & 50.1 \\ 
\ \ +ReSCORE & 43.8 & 58.0 & 47.3 & 63.3 & 77.2 \\ 
\cdashline{1-6}[2pt/1pt]
\noalign{\vskip 2pt}
Our Baseline & 39.4 & 52.3 & 44.8 & 47.5 & 47.5 \\
\ \ +ReSCORE & 47.2 & 59.3 & 46.6 & 69.3 & 72.4 \\ 
\midrule
\multicolumn{6}{c}{\textbf{2WikiMHQA}} \\
\midrule
$\text{Self-RAG}^{*}$  & 3.0 & 19.1 & 26.3 & 27.1 & 27.1 \\ 
\ \ +ReSCORE & 5.6 & 21.2 & 25.9 & 28.4 & 32.8 \\
\cdashline{1-6}[2pt/1pt]
\noalign{\vskip 2pt}
FLARE & 23.2 & 35.0 & 32.5 & 42.9 & 42.9 \\ 
\ \ +ReSCORE & 26.5 & 38.0 & 33.2 & 45.6 & 45.6 \\ 
\cdashline{1-6}[2pt/1pt]
\noalign{\vskip 2pt}
Adaptive-Note  & 35.7 & 46.1 & 45.7 & 59.2 & 59.2 \\ 
\ \ +ReSCORE & 37.4 & 49.3 & 49.8 & 63.2 & 67.5 \\ 
\cdashline{1-6}[2pt/1pt]
\noalign{\vskip 2pt}
Our Baseline & 32.8 & 41.6 & 45.7 & 56.9 & 56.9 \\
\ \ +ReSCORE & 50.0 & 59.7 & 51.2 & 81.2 & 88.0 \\ 
\bottomrule
\end{tabular}
\caption{\textbf{Effects of ReSCORE with various iterative RAG systems on three MHQA benchmarks.} 
All methods are re-implemented using Llama 3.1 and Contriever, except for Self-RAG, which uses Llama-2-7B model from the original study. 
All hyperparameters for the baselines are taken from the original paper and code, as detailed in \cref{sec:hyperparameters}.}
\label{tab:base_model_comparison}
\end{table}

%% file: tables/02_relevance_table.tex
\begin{table}[t]
\small 
\centering
\begin{tabular}{lccccc}
\toprule
\textbf{Pseudo-GT} & \textbf{EM} & \textbf{F1} & \textbf{R@2} & \textbf{R@4} & \textbf{R@8}  \\ 
\midrule
\multicolumn{6}{c}{\textbf{MuSiQue}} \\
\midrule
None & 15.2 & 23.8 & 32.7 & 40.1 & 47.1   \\
\cdashline{1-6}[2pt/1pt]
\noalign{\vskip 2pt}
$P_{\text{LM}}(q \mid d)$ & 15.9 & 25.9 & 34.6 & 41.1 & 47.9   \\
$P_{\text{LM}}(a \mid q, d)$ & 5.8 & 12.3 & 28.9 & 35.1 & 41.4   \\
{$P_{\text{LM}}(q, a \mid d)$} & \textbf{16.4} & \textbf{26.3} & \textbf{42.7} & \textbf{50.3} & \textbf{55.7}   \\ 
\midrule
\multicolumn{6}{c}{\textbf{HotpotQA}} \\
\midrule
None & 39.4 & 52.3 & 49.4 & 56.5 & 61.7  \\
\cdashline{1-6}[2pt/1pt]
\noalign{\vskip 2pt}
$P_{\text{LM}}(q \mid d)$ & 42.0 & 53.9 & 55.2 & 62.4 & 65.9   \\
$P_{\text{LM}}(a \mid q, d)$ & 19.2 & 26.4 & 27.5 & 34.4 & 42.8   \\
{$P_{\text{LM}}(q, a \mid d)$} & \textbf{43.6} & \textbf{56.4} & \textbf{58.1} & \textbf{64.6} & \textbf{68.3}   \\ 
\midrule
\multicolumn{6}{c}{\textbf{2WikiMHQA}} \\
\midrule 
None & 32.8 & 41.6 & 46.4 & 54.3 & 58.9  \\
\cdashline{1-6}[2pt/1pt]
\noalign{\vskip 2pt}
$P_{\text{LM}}(q \mid d)$ & 39.2 & 47.9 & 50.8 & 59.1 & 63.2   \\
$P_{\text{LM}}(a \mid q, d)$ & 18.8 & 26.5 & 26.1 & 33.3 & 41.9   \\
{$P_{\text{LM}}(q, a \mid d)$} & \textbf{41.4} & \textbf{51.7} & \textbf{53.7} & \textbf{63.0} & \textbf{67.1}   \\ 
\bottomrule
\end{tabular}
\caption{\textbf{Comparisons of Different Pseudo-GT Labels on Document Reranking.} Recall@$k$ (R@$k$) was computed after retrieving 100 documents with Contriever and re-ranking them using the given pseudo-GT label for questions in the validation set.
EM/F1 was computed in the same setting on the test set.
}
\label{tab:document_relevance}
\end{table}

%% file: tables/04_training_signal_table.tex
\begin{table}[t]
\centering
\small
\begin{tabular}{lccccc}
\toprule
 & \multicolumn{2}{c}{\textbf{QA}} & \multicolumn{3}{c}{\textbf{$\text{MHR}_{i}@\text{8}$}} \\
\textbf{Label} & \textbf{EM} & \textbf{F1} & $i=1$ & $i=2$ & $i=\eta_n$ \\
\midrule
\multicolumn{6}{c}{\textbf{MuSiQue}} \\
\midrule
None         & 15.2 & 23.8 & 44.9 & 51.6 & 51.6 \\
GT      & 15.8 & 24.9 & 46.7 & 54.8 & 54.8 \\
Pseudo-GT        & 23.4 & 32.7 & 46.8 & 63.0 & 65.2 \\
\midrule
\multicolumn{6}{c}{\textbf{HotpotQA}} \\
\midrule
None         & 39.4 & 52.3 & 44.8 & 47.5 & 47.5 \\
GT     & 45.2 & 55.8 & 48.7 & 52.7 & 52.7 \\
Pseudo-GT        & 47.2 & 59.3 & 46.6 & 69.3 & 72.4 \\
\midrule
\multicolumn{6}{c}{\textbf{2WikiMHQA}} \\
\midrule
None         & 32.8 & 41.6 & 45.7 & 56.9 & 56.9 \\
GT     & 37.1 & 46.2 & 48.5 & 61.7 & 61.7 \\
Pseudo-GT        & 50.0 & 59.7 & 51.2 & 81.2 & 88.0 \\
\bottomrule
\end{tabular}
\caption{\textbf{Comparisons of Different Labels for fine-tuning retrievers on three MHQA benchmarks.} 
None denotes no label, which means the baseline model without fine-tuning.
GT is a binary label denoting whether a document is relevant to a given question or not.
Pseudo-GT is the labels used within ReSCORE.
}
\label{tab:multistep_ablation}
\end{table}

%% file: figures/03_fig.tex
\begin{figure*}[ht]
    \centering
    \begin{subfigure}[t]{0.31\textwidth}
        \centering
        \includegraphics[width=\textwidth, trim=0cm 0cm 0cm 0cm, clip]{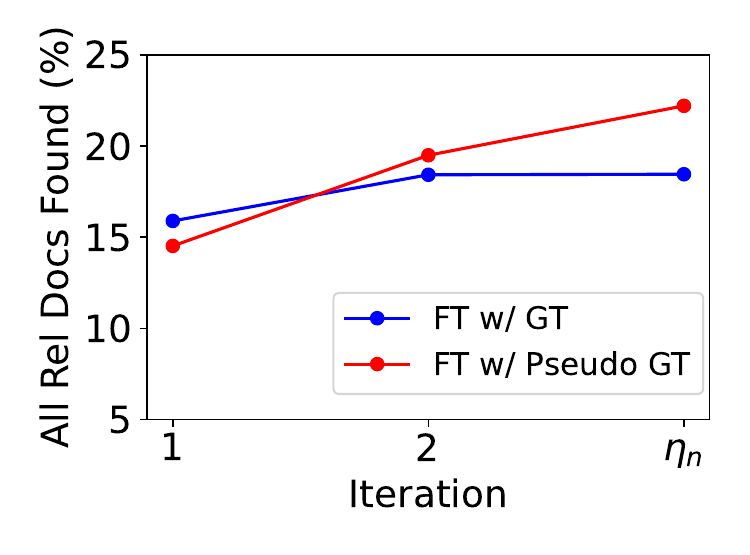}
        \vspace{-0.7cm}
        \caption{\ MuSiQue Dataset}
        \label{fig:graph1}
    \end{subfigure}
    \hspace{0.01\textwidth}
    \begin{subfigure}[t]{0.31\textwidth}
        \centering
        \includegraphics[width=\textwidth, trim=0cm 0cm 0cm 0cm, clip]{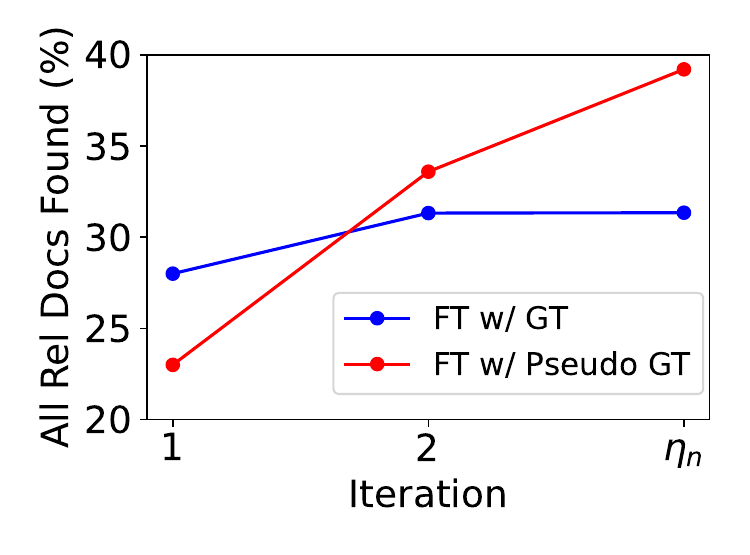}
        \vspace{-0.7cm}
        \caption{\ HotpotQA Dataset}
        \label{fig:graph2}
    \end{subfigure}
    \hspace{0.01\textwidth}
    \begin{subfigure}[t]{0.31\textwidth}
        \centering
        \includegraphics[width=\textwidth, trim=0cm 0cm 0cm 0cm, clip]{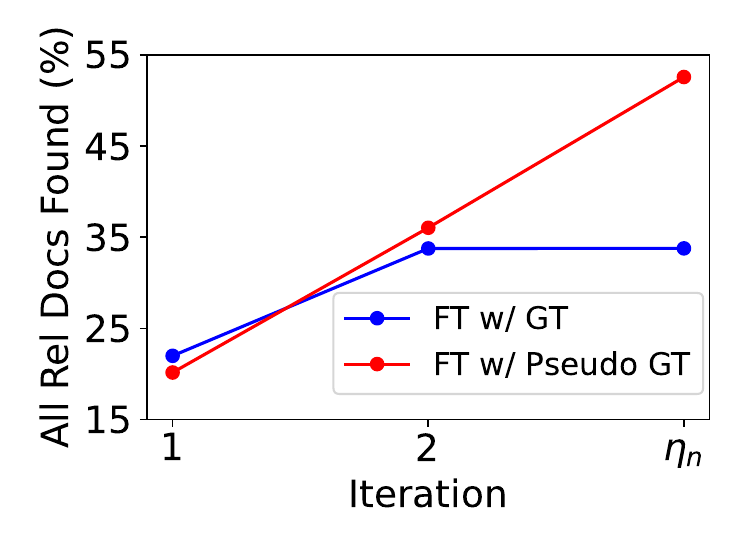}
        \vspace{-0.7cm}
        \caption{\ 2WikiMHQA Dataset}
        \label{fig:graph3}
    \end{subfigure}
    \caption{\textbf{Comparison of GT and Pseudo-GT Labels on All Relevant Document Retrieval.} The y-axis shows the proportion of questions for which \texttt{all} relevant documents were found, which are all needed to correctly answer a given complex question.
    Pseudo-GT labels lead to improved performance as the number of iterations increases.
    }
    \label{fig:graph}
\end{figure*}

%% file: tables/05_rewriting_table.tex
\begin{table}[t]
\centering
\small
\begin{tabular}{lccccc}
\toprule
\textbf{Reformulation} & \multicolumn{2}{c}{\textbf{QA}} & \multicolumn{3}{c}{\textbf{$\text{MHR}_{i}@\text{8}$}} \\
\textbf{Method} & \textbf{EM} & \textbf{F1} & $i=1$ & $i=2$ & $i=\eta_n$ \\
\midrule 
\multicolumn{6}{c}{\textbf{MuSiQue}} \\
\midrule 
None         & 10.8 & 17.8 & 44.7 & 45.4 & 47.4 \\ 
LLM-rewrite  & 21.2 & 30.5 & 45.1 & 56.7 & 63.7 \\ 
Thought-concat      & 23.4 & 32.7 & 46.8 & 63.0 & 65.2 \\
\midrule 
\multicolumn{6}{c}{\textbf{HotpotQA}} \\
\midrule
None         & 29.4 & 41.1 & 42.8 & 43.6 & 43.8 \\ 
LLM-rewrite  & 44.2 & 57.4 & 41.9 & 54.8 & 64.7 \\ 
Thought-concat     & 47.2 & 59.3 & 46.6 & 69.3 & 72.4 \\
\midrule 
\multicolumn{6}{c}{\textbf{2WikiMHQA}} \\
\midrule 
None         & 35.6 & 44.7 & 48.6 & 49.7 & 49.8 \\ 
LLM-rewrite  & 51.7 & 60.1 & 50.0 & 86.0 & 89.5 \\ 
Thought-concat     & 50.0 & 59.7 & 51.2 & 81.2 & 88.0 \\
\bottomrule
\end{tabular}
\caption{\textbf{Effect of Query Reformulation Methods on MHQA.} 
We compare three methods: (1) no rewriting (None), (2) LLM-based query rewriting using retrieved documents (LLM-rewrite), and (3) concatenating summarized \textit{thoughts} to the original query for retrieval (Thought-concat).
}
\label{tab:rewriting}
\end{table}

%% file: src/05_conclusion.tex
\section{Conclusion}
% short version
% We presented ReSCORE, a novel method for training dense retrievers for MHQA without labeled documents and incorporated it into IQATR to achieve the new SOTA on MHQA.
% We also employed it in existing MHQA systems to improve the performance, showcasing its broad applicability to various iterative RAG frameworks for MHQA.

% longer version
In this paper, 
we presented ReSCORE, a novel method for training dense retrievers for MHQA without documents labeled with their relevance to respective queries. 
To demonstrate the efficacy of ReSCORE,
we incorporated it into an iterative RAG framework, IQATR, to achieve the new SOTA on MHQA.
We also employed it in existing MHQA systems to improve the performance, showcasing its broad applicability to various iterative RAG frameworks for MHQA.
In addition, 
we conducted additional experiments to analyze various query reformulation methods and pseudo-GT labels to be used as fine-tuning signals for retriever training.
We expect our in-depth analysis to provide deeper insights into ReSCORE and help devise ways to improve on this label-free retriever training method.

%% file: src/06_appendix.tex
\appendix
\label{sec:appendix}
\section{Hyperparameters}
\label{sec:hyperparameters}
Hyperparameters are directly adopted from the original papers or their accompanying codes, as detailed in the Tab~\ref{tab:hyperparams}. 
It is important to note that these hyperparameters result from an engineering process specifically tailored for each method, and therefore we have adopted them as they are to ensure consistency. 
When applying our method, ReSCORE, in Tab~\ref{tab:base_model_comparison}, we adhered to the same hyperparameters to ensure fair comparisons across all methods. 
For prompts, we refer to those detailed in each corresponding paper.
The terms listed in Table~\ref{tab:hyperparams} are explained as follows:  
\textbf{Max Tokens} refers to the maximum number of tokens allowed in the generated output, limiting its length.  
\textbf{$T$} represents the model's generation temperature, which is set to 0. to ensure consistent outputs.  
\textbf{Top-$k$} indicates the number of documents considered for answering at each iteration.  
\textbf{Max Step} defines the maximum number of iterations the model can perform.  
\textbf{Max Fail} specifies the maximum number of retries allowed in case an iteration fails. 
For training the baselines with ReSCORE, we use the same number of documents, \textbf{$M=32$}, for the distribution as used in IQATR across all experiments.

\section{Details of the ReSCORE Framework}
In this section, we provide a detailed explanation of the prompts utilized in our framework, outlining their roles and usage across different components.

The following prompts are employed in the framework:
\begin{itemize}[itemsep=2pt, topsep=0pt, parsep=0pt, partopsep=0pt]
    \item \textit{Answer Generation Prompt} (Appendix~\ref{sec:answer_generation}): Used to either generate answers or explicitly indicate when the model does not know the answer, clarifying whether to continue the iteration or stop.
    \item \textit{Thought Generation Prompt} (Appendix~\ref{sec:thought_generation}): Guides the extraction of relevant information from retrieved documents by summarizing and preventing context overflow, ensuring the model stays within the context limit.
    \item \textit{Question Rewriting Prompt} (Appendix~\ref{sec:question_rewriting}): Employed specifically for LLM-based question rewriting tasks, as illustrated in Table~\ref{tab:rewriting}.
\end{itemize}

\vspace{0.2em}

For document relevance evaluation, we explore three key prompts:
\begin{itemize}[itemsep=2pt, topsep=0pt, parsep=0pt, partopsep=0pt]
    \item The $P_{\text{LM}}(a \mid q, d)$ prompt (Appendix~\ref{sec:promptaqd}), which evaluates the likelihood of an answer $a$ given a question $q$ and document $d$.
    \item The $P_{\text{LM}}(q \mid d)$ prompt (Appendix~\ref{sec:q}), which assesses the relevance of a question $q$ to the document $d$.
    \item The $P_{\text{LM}}(q, a \mid d)$ prompt (Appendix~\ref{sec:q&a}), which jointly considers the likelihood of a question-answer pair $(q, a)$ given the document $d$.
\end{itemize}

Among these, the $P_{\text{LM}}(q, a \mid d)$ prompt serves as the default pseudo-GT generation mechanism in the ReSCORE framework. 

\input{tables/06_hyperparameter_table}
\input{tables/07_stats}
\input{tables/08_extended_baseline}

\section{Statistical Significance}
\label{sec:stats}
% \phseo{Reference to the table}
In this section, we assess the statistical significance of the results shown in Tab.\ref{tab:sota} by performing Student’s t-tests, summarized in Tab.\ref{tab:significance_results}.
We compare our method against Adaptive-RAG~\cite{jeong2024adaptive}, IRCoT~\cite{trivedi2022interleaving}, and Adaptive-Note~\cite{wang2024retriever} over 10 independent runs with different random seeds.
Our approach consistently achieves statistically significant improvements across all evaluated benchmarks (p-value < 0.05 for all comparisons).
Specifically, our model outperforms these baselines on the MuSiQue, HotpotQA, and 2WikiMHQA datasets, with significant gains in both EM and F1 scores.
Even in the least favorable case—MuSiQue EM—our approach maintains significance (p = 0.045), with an average 0.92-point improvement over Adaptive-RAG.
Even in the least favorable case—MuSiQue EM—our approach maintains significance (p = 0.045), with an average 0.92-point improvement over Adaptive-RAG.
All other comparisons yield even stronger statistical significance (p $\ll$ 0.05), further confirming the robustness of our method.

Additionally, the significance of ReSCORE is evident in several key aspects. 
First, it achieves consistent performance across all three datasets without requiring benchmark-specific hyperparameter tuning. 
In contrast, IRCoT and Adaptive-RAG adjust their hyperparameters and few-shot prompts, even leveraging GT document annotations.
Moreover, Adaptive-Note incorporates GPT-3.5, which has stronger reasoning capabilities compared to Llama, the model used in this work.
Furthermore, as shown in Tab~\ref{tab:base_model_comparison},
ReSCORE consistently enhances the performance of Adaptive-Note in both retrieval and MHQA metrics across all three benchmarks.
These findings underscore the robustness and effectiveness of the proposed method.

\section{Extended Comparison to Baselines}
% Moved this here due to ACL style format
\label{sec:extended}
The comparison to baselines in Tab~\ref{tab:sota} is extended to include additional iterative RAG frameworks in this section. Self-Ask~\cite{press-etal-2023-measuring} and SearChain~\cite{10.1145/3589334.3645363}, both of which leverage GPT-3.5. 
These methods employ a step-by-step reasoning process, iteratively refining retrieved information while generating the final response.
To further quantify performance, the cover-EM (cEM) metric from SearChain is introduced to IQATR. 
cEM assigns a score of 1 if the exact answer tokens appear anywhere in the LLM-generated output. 
As shown in Tab~\ref{tab:comparisons}, IQATR outperforms Self-Ask by an average of 7.1 on EM and F1, and also surpasses SearChain by an average of 8.8 on cEM, despite both baselines using GPT-3.5 while our model uses Llama-8B.

\clearpage
\clearpage

\section{Prompts}
\label{sec:a}
\subsection{Answer Generation Prompt}
\label{sec:answer_generation}
\begin{tcolorbox}[colframe=gray!50!black, colback=gray!10!white, title=Answer Generation Prompt, width=\textwidth, sharp corners=southwest]
\begin{verbatim}
<|start_header_id|>system<|end_header_id|>

You will receive three inputs: 'documents', 'a question', and 'hints'.  
Your task is to answer the given question.

Instructions:
 - Carefully read the documents and hints. 
 - If you know the answer to the question confidently, generate an answer, 
 using documents and hints provided.
 - If you don't know, generate "Unknown".

Format:
 - Return a JSON object formatted as follows: {{"answer": "Your Response"}}
 - Your response should be concise 'short-answer' 
 without any explanation or "Unknown".
 - Ensure the entire response is on a single line without placeholder variables.

You are a helpful assistant.<|eot_id|><|start_header_id|>user<|end_header_id|>

Documents:
{documents}

Question: 
{question}

Hints:
{hints}
<|eot_id|><|start_header_id|>assistant<|end_header_id|>
\end{verbatim}
\end{tcolorbox}

\clearpage
\subsection{Thought Generation Prompt}
\label{sec:thought_generation}
\begin{tcolorbox}[colframe=gray!50!black, colback=gray!10!white, title=Thought Generation Prompt, width=\textwidth, sharp corners=southwest]
\begin{verbatim}
<|start_header_id|>system<|end_header_id|>

You will receive three inputs: 'documents', 'a question', and 'hints'.
Your task is to provide a hint that aids answering the given question.

Instructions:
 - Carefully read the documents and hints. 
 - Generate a hint containing partial information relevant to the question, 
 using documents and hints provided.

Format:
 - Return a JSON object in this format: {{"hint": "Your response"}}
 - Your response should be concise 'one-sentence hint'.
 - Ensure the entire response is on a single line without placeholder variables.

You are a helpful assistant.<|eot_id|><|start_header_id|>user<|end_header_id|>

Documents:
{documents}

Question: 
{question}

Hints:
{hints}
<|eot_id|><|start_header_id|>assistant<|end_header_id|>
\end{verbatim}
\end{tcolorbox}
\clearpage
\subsection{Question Rewriting Prompt}
\label{sec:question_rewriting}
\begin{tcolorbox}[colframe=gray!50!black, colback=gray!10!white, title=Question Rewriting Prompt, width=\textwidth, sharp corners=southwest]
\begin{Verbatim}
<|start_header_id|>system<|end_header_id|>

You will receive two inputs: 'documents', and a 'question'.  
Your task is to create a new question that asks for additional documents 
or information required to comprehensively answer the original question.

Instructions: 
- Analyze the provided documents and identify any missing information, 
entities, or relationships needed to fully answer the original question.
- Formulate a new question that explicitly asks for the missing 
information or documents needed.
- Ensure that the new question maintains the original context and 
scope of the original question.
- Focus on identifying gaps in entities (people, places, events) 
or specific details that are absent from the provided documents 
but are necessary to answer the original question.

Format: 
- Return a JSON object formatted as follows: {{"question": "<Your Response>"}}
- Ensure the entire response is on a single line 
without placeholder variables or assumptions.

You are a helpful assistant.<|eot_id|><|start_header_id|>user<|end_header_id|> 

{documents}

Question: {question}<|eot_id|><|start_header_id|>assistant<|end_header_id|>
\end{Verbatim}
\end{tcolorbox}

\clearpage

\subsection{$P_{\text{LM}}(a \mid q,d)$ Prompt}
\label{sec:promptaqd}
\begin{tcolorbox}[colframe=gray!50!black, colback=gray!10!white, title=Condition Prompt, width=\textwidth, sharp corners=southwest]
\begin{verbatim}
<|start_header_id|>system<|end_header_id|>

Your task is to answer the given question using the given document(s).

Instructions:
 - Carefully read the provided document(s).
 - Answer the question using the given document(s).

Format:
 - Return a JSON object formatted as follows: 
{{
    "answer": "The short-form answer to the question."
}}
 - Your response should be concise 'short-answer'.
 - Ensure the entire response is on a single line without placeholder variables.

You are a helpful assistant.<|eot_id|><|start_header_id|>user<|end_header_id|>

Document(s):
{documents}

Question: 
{question}
<|eot_id|><|start_header_id|>assistant<|end_header_id|>
\end{verbatim}
\end{tcolorbox}
\begin{tcolorbox}[colframe=gray!50!black, colback=gray!10!white, title=Prediction Prompt, width=\textwidth, sharp corners=southwest]
\begin{verbatim}
{{
    "answer": "{answer}"
}}
\end{verbatim}
\end{tcolorbox}
\clearpage
\subsection{$P_{\text{LM}}(q \mid d)$ Prompt}
\label{sec:q}
\begin{tcolorbox}[colframe=gray!50!black, colback=gray!10!white, title=Condition Prompt, width=\textwidth, sharp corners=southwest]
\begin{verbatim}
<|start_header_id|>system<|end_header_id|>

Your task is to generate a question using the given document(s).

Instructions:
 - Carefully read the provided document(s).
 - Create a question that can be answered using the given document(s).
 - Use information from one or more documents, but ensure that the answer is concise 
 and directly supported by the content.

Format:
 - Return a JSON object formatted as follows: 
{{
    "question": "Your generated question based on the documents.",
}}
 - Make sure the question is on-topic.
 - Ensure the entire response is on a single line without placeholder variables.

You are a helpful assistant.<|eot_id|><|start_header_id|>user<|end_header_id|>

Document(s):
{documents}
<|eot_id|><|start_header_id|>assistant<|end_header_id|>
\end{verbatim}
\end{tcolorbox}
\begin{tcolorbox}[colframe=gray!50!black, colback=gray!10!white, title=Prediction Prompt, width=\textwidth, sharp corners=southwest]
\begin{verbatim}
{{
    "question": "{question}"
}}
\end{verbatim}
\end{tcolorbox}
\clearpage

\subsection{$P_{\text{LM}}(q, a \mid d)$ Prompt}
\label{sec:q&a}
\begin{tcolorbox}[colframe=gray!50!black, colback=gray!10!white, title=Condition Prompt, width=\textwidth, sharp corners=southwest]
\begin{verbatim}
<|start_header_id|>system<|end_header_id|>

Your task is to generate a question-answer pair using the given document(s).

Instructions:
 - Carefully read the provided document(s).
 - Create a question that can be answered using the given document(s).
 - Use information from one or more documents, but ensure that the answer is 
 concise and directly supported by the content.

Format:
 - Return a JSON object formatted as follows: 
{{
    "question": "Your generated question based on the documents.",
    "answer": "The short-form answer to the question."
}}
 - Make sure the question is on-topic and the answer is concise.
 - Ensure the entire response is on a single line without placeholder variables.

You are a helpful assistant.<|eot_id|><|start_header_id|>user<|end_header_id>

Document(s):
{documents}
<|eot_id|><|start_header_id|>assistant<|end_header_id>
\end{verbatim}
\end{tcolorbox}
\begin{tcolorbox}[colframe=gray!50!black, colback=gray!10!white, title=Prediction Prompt, width=\textwidth, sharp corners=southwest]
\begin{verbatim}
{{
    "question": "{question}",
    "answer": "{answer}"
}}
\end{verbatim}
\end{tcolorbox}

%% file: tables/06_hyperparameter_table.tex
\begin{table}[t]
    \small
    \centering
    \begin{tabular}{lccccc}
        \toprule
         & \textbf{Max} &  & \textbf{Top} & \textbf{Max} & \textbf{Max} \\
         & \textbf{Tokens} & \textbf{$T$} & \textbf{$k$} & \textbf{Step} & \textbf{Fail} \\
        \midrule
        FLARE & 256 & 0. & 2 & 7 & - \\
        Self-RAG & 50 & 0. & 5 & 10 & - \\
        Adaptive-Note & 1280 & 0. & 8 & 3 & 2 \\
        Ours & 64 & 0. & 8 & 6 & - \\
        \bottomrule
    \end{tabular}
    \caption{\textbf{Hyperparameters used for reproducing each method}. The hyperparameters and prompts are adopted directly from the original papers or their accompanying code to ensure consistency.}
    \label{tab:hyperparams}
\end{table}

%% file: tables/07_stats.tex
\begin{table*}[t]
    \centering
    \small
    \scalebox{1.0}{
    \begin{tabular}{lccccccc}
        \toprule
        \textbf{ } & \textbf{ } 
        & \multicolumn{2}{c}{\textbf{MuSiQue}} 
        & \multicolumn{2}{c}{\textbf{HotpotQA}} 
        & \multicolumn{2}{c}{\textbf{2WikiMHQA}} \\
        \cmidrule(lr){3-4} \cmidrule(lr){5-6} \cmidrule(lr){7-8}
         &  
         & \textbf{EM} & \textbf{F1} 
         & \textbf{EM} & \textbf{F1} 
         & \textbf{EM} & \textbf{F1} \\
        \midrule
        \multirow{2}{*}{AdaptiveNote (GPT-3.5+BM25)} 
            & $p$-value
                & 1.4e-18 & 2.5e-17 
                & 3.0e-05 & 8.8e-3 
                & 1.7e-14 & 1.1e-14 \\
            & $\Delta$ 
                & +9.58 & +8.16 
                & +1.28 & +0.77 
                & +6.36 & +5.51 \\
        \noalign{\vskip 1.5pt}
        \cdashline{1-8}[2pt/1pt]
        \noalign{\vskip 2pt}
        \multirow{2}{*}{IRCoT (Flan-T5-XL+BM25)} 
            & $p$-value 
                & 2.2e-3 & 5.9e-3 
                & 3.9e-13 & 1.2e-08 
                & 3.5e-2 & 7.5e-18 \\
            & $\Delta$ 
                & +0.86 & +0.72 
                & +2.74 & +2.67 
                & +0.31 & +5.09 \\
        \noalign{\vskip 1.5pt}
        \cdashline{1-8}[2pt/1pt]
        \noalign{\vskip 2pt}
        \multirow{2}{*}{Adaptive-RAG (Flan-T5-XL+BM25)} 
            & $p$-value 
                & 4.5e-2 & 9.8e-3 
                & 1.3e-18 & 2.9e-18 
                & 2.5e-23 & 1.4e-22 \\
            & $\Delta$ 
                & +0.98 & +0.64 
                & +5.09 & +5.35 
                & +9.40 & +10.08 \\
        \bottomrule
    \end{tabular}
    }
    \caption{
    \textbf{Student's t-test Results for Ours vs. Baselines.}
    $p$-values from a two-tailed Student's t-test over 10 random seeds show all differences are statistically significant (< 0.05).
    $\Delta$ indicates the average performance gap: \textit{Ours – Baseline}, with positive $\Delta$ meaning \textit{Ours} performed better.
    }
    \label{tab:significance_results}
\end{table*}

%% file: tables/08_extended_baseline.tex
\begin{table*}[t]
    \small
    \centering
    \begin{tabular}{lccccccccc}
        \toprule
        & \multicolumn{3}{c}{\textbf{MuSiQue}} & \multicolumn{3}{c}{\textbf{HotpotQA}} & \multicolumn{3}{c}{\textbf{2WikiMHQA}} \\
        \cmidrule(lr){2-4} \cmidrule(lr){5-7} \cmidrule(lr){8-10}
        & \textbf{cEM} & \textbf{EM} & \textbf{F1} & \textbf{cEM} & \textbf{EM} & \textbf{F1} & \textbf{cEM} & \textbf{EM} & \textbf{F1} \\
        \midrule
        Self-Ask (GPT-3.5) $\ddagger\ddagger$ & - & 13.8 & 27.0 & - & - & - & - & 30.0 & 36.1 \\
        Self-Ask (GPT-3.5+Google Search API)$\ddagger\ddagger$ & - & 15.2 & 27.2 & - & - & - & - & 40.1 & 52.6 \\
        SearChain (GPT-3.5+ColBERT)$\ddagger\ddagger$ & 17.1 & - & - & 56.9 & - & - & 46.3 & - & - \\
        \noalign{\vskip 1.5pt}
        \cdashline{1-10}[2pt/1pt]
        \noalign{\vskip 2pt}
        Ours (Llama-8B+Contriever+ReSCORE) & \textbf{30.4} & \textbf{23.4} & \textbf{32.7} & \textbf{59.6} & \textbf{47.2} & \textbf{59.3} & \textbf{57.0} & \textbf{50.0} & \textbf{59.7} \\
        \bottomrule
    \end{tabular}
    \caption{\textbf{Extended Comparison with Iterative Frameworks.} cEM is a metric that assigns a score of 1 if the exact answer tokens appear anywhere in the LLM-generated output.
    $\ddagger\ddagger$ Scores are sourced from the original papers.}
    \label{tab:comparisons}
\end{table*}